\documentclass[final]{nesy2025} % Uncomment to include author names

\usepackage{longtable}% for long tables

\usepackage{booktabs}
\usepackage[load-configurations=version-1]{siunitx} % newer version
\usepackage{siunitx}

\usepackage{algorithm}
\usepackage{algpseudocode}
\usepackage{array}
\usepackage{booktabs}
\usepackage{amsmath}
\usepackage[a4paper,margin=1in]{geometry}

\usepackage{enumitem}
\usepackage{amsmath}
\usepackage{amssymb}

\usepackage{color}
\usepackage{listings}
\usepackage{xcolor}

\definecolor{background}{rgb}{0.95,0.95,0.95}
\definecolor{numb}{rgb}{0.5,0.5,0.5}
\definecolor{punct}{rgb}{0.3,0.3,0.3}
\definecolor{delim}{rgb}{0.6,0.2,0.2}

\lstdefinelanguage{json}{
    basicstyle=\footnotesize\ttfamily,
    numbers=left,
    numberstyle=\tiny,
    stepnumber=1,
    numbersep=4pt,
    showstringspaces=false,
    breaklines=true,
    frame=lines,
    backgroundcolor=\color{background},
    literate=
     *{0}{{{\color{numb}0}}}{1}
      {1}{{{\color{numb}1}}}{1}
      {2}{{{\color{numb}2}}}{1}
      {3}{{{\color{numb}3}}}{1}
      {4}{{{\color{numb}4}}}{1}
      {5}{{{\color{numb}5}}}{1}
      {6}{{{\color{numb}6}}}{1}
      {7}{{{\color{numb}7}}}{1}
      {8}{{{\color{numb}8}}}{1}
      {9}{{{\color{numb}9}}}{1}
      {:}{{{\color{punct}{:}}}}{1}
      {,}{{{\color{punct}{,}}}}{1}
      {\{}{{{\color{delim}{\{}}}}{1}
      {\}}{{{\color{delim}{\}}}}}{1}
      {[}{{{\color{delim}{[}}}}{1}
      {]}{{{\color{delim}{]}}}}{1},
}
\lstdefinelanguage{Isar}{
    basicstyle=\footnotesize\ttfamily,
    numbers=left,
    numberstyle=\tiny,
    stepnumber=1,
    numbersep=4pt,
    showstringspaces=false,
    breaklines=true,
    frame=lines,
    backgroundcolor=\color{background},
    literate=
     *{0}{{{\color{numb}0}}}{1}
      {1}{{{\color{numb}1}}}{1}
      {2}{{{\color{numb}2}}}{1}
      {3}{{{\color{numb}3}}}{1}
      {4}{{{\color{numb}4}}}{1}
      {5}{{{\color{numb}5}}}{1}
      {6}{{{\color{numb}6}}}{1}
      {7}{{{\color{numb}7}}}{1}
      {8}{{{\color{numb}8}}}{1}
      {9}{{{\color{numb}9}}}{1}
      {:}{{{\color{punct}{:}}}}{1}
      {,}{{{\color{punct}{,}}}}{1}
      {\{}{{{\color{delim}{\{}}}}{1}
      {\}}{{{\color{delim}{\}}}}}{1}
      {[}{{{\color{delim}{[}}}}{1}
      {]}{{{\color{delim}{]}}}}{1}
      {->}{$\rightarrow$}{2}
      {∈}{$\in$}{2}
      {∧}{$\wedge$}{2},
}

\theorembodyfont{\upshape}
\theoremheaderfont{\scshape}
\theorempostheader{:}
\theoremsep{\newline}

\title[ProofSeek]{Neural Theorem Proving: Generating and Structuring Proofs for Formal Verification}

 \author{\Name{Balaji Rao} \Email{brao@stevens.edu}\\
   \Name{William Eiers} \Email{weiers@stevens.edu}\\
   \Name{Carlo Lipizzi} \Email{clipizzi@stevens.edu}\\
   \addr 1 Castle Point Terrace, Hoboken, NJ 07030}

\begin{document}

\maketitle

\begin{abstract}
Formally verifying properties of software code has been a highly desirable task, especially with the emergence of LLM-generated code. In the same vein, they provide an interesting avenue for the exploration of formal verification and mechanistic interpretability. Since the introduction of code-specific models, despite their successes in generating code in Lean4 and Isabelle, the task of generalized theorem proving still remains far from being fully solved and will be a benchmark for reasoning capability in LLMs. In this work, we introduce a framework that generates whole proofs in a formal language to be used within systems that utilize the power of built-in tactics and off-the-shelf automated theorem provers. Our framework includes 3 components: generating natural language statements of the code to be verified, an LLM that generates formal proofs for the given statement, and a module employing heuristics for building the final proof. To train the LLM, we employ a 2-stage fine-tuning process, where we first use SFT-based training to enable the model to generate syntactically correct Isabelle code and then RL-based training that encourages the model to generate proofs verified by a theorem prover. We validate our framework using the miniF2F-test benchmark and the Isabelle proof assistant and design a use case to verify the correctness of the AWS S3 bucket access policy code. We also curate a dataset based on the FVEL\textsubscript{\textnormal{ER}} dataset for future training tasks\footnote{The code, dataset and training scripts are available at: \url{https://github.com/kings-crown/ProofSeek}}.
\end{abstract}

% \keywords{Neural Theorem Proving, Large Language Models, Reinforcement Learning, Formal Methods, Automated Theorem Proving}

\section{Introduction}

Recent advances in language models has revolutionized the approach to mathematical reasoning in artificial intelligence. Language models, and in particular large language models (LLMs), have made significant advances in the field of general theorem proving~\citep{guo2025deepseek, azerbayev2023llemma}. Consequently, formal theorem proving using large language models has recently garnered renewed attention. 
Formal theorem proving lies at the intersection of mathematics and computer science, where mathematical statements modeling the interaction of computer systems are derived and translated into a formal language which is used to prove the correctness of programs. While formal theorem proving is effective at producing high quality code that provides correctness~\citep{bibel2013automated} it is often a laborious process and requires an intimate level of domain expertise in order to correctly model the computer program into a mathematical statement. Moreover, the costs associated with manual verification can be prohibitively expensive and potentially yield unwieldy proofs which are far more complex than the code being verified.

% As a consequence formal theorem proving, has recently garnered renewed attention. It is field that lies at the intersection of mathematics and computer science. It is the process of writing mathematic statements into a formal language to be checked for correctness by a program, and while offering robustness for the statements proved, is often labor-intensive and requires a high level of expertise. It is effective at producing high quality code that provides correctness~\cite{bibel2013automated}. However, the costs associated with manual verification is high and the proofs themselves end up being many times longer than the code being verified.

There have been a number of approaches leveraging machine learning in automated theorem proving which focus on tasks such as premise selection~\citep{irving2016deepmath} and proof search~\citep{loos2017deep}. More recently, with advancements in LLMs' mathematical reasoning abilities~\citep{azerbayev2023llemma,shao2024deepseekmath}, there have been focused efforts on using LLMs for automated proof synthesis. Two main paradigms towards automated theorem proving have emerged: one that generates the whole proof and the other that generates only the next proof step. However, both systems suffer from several shortcomings. Whole proof generators are incapable of utilizing previously proved lemmas. On the other hand, proof step generators are not as scalable during training and inference~\citep{xin2024deepseek}. 

While there are a number of studies that attempt to address the shortcoming of each of these paradigms \citep{wang2023lego,zheng2023lyra,dong2024formal}, they all concentrate on silos of mathematical processes and build systems to perform better on benchmarks. They do not look at systems built for purposes beyond the mathematical community.

In this paper, we introduce a framework for generalized theorem proving to bridge the gap left by existing neural theorem proving approaches which mainly focus on achieving high proof success rate on standardized benchmarks. Our framework consists of three core modules: the first generates natural language statements of the code, policy, or statement to be verified; the second trains an LLM which generates formal proofs from the natural language statement; the final module employs heuristics from the ProofAug approach~\citep{liu2025efficient} for building the final proof which can be checked by the proof assistants such as Isabelle. We employ a two-stage fine-tuning process to train the LLM, where we leverage SFT-based training to enable the model to generate syntactically valid Isabelle code, and RL-based training to encourage the model to generate semantically valid Isabelle code. The main contributions of this work are as follows:

\begin{itemize}[leftmargin=1.5em,itemsep=-.7pt,parsep=0.5pt,topsep=0pt]
    \item We introduce a framework for generalized theorem proving that enables the verification of natural language statements, including code and security policies.
    \item We curate a dataset to fine-tune language models for formal theorem proving, improving their reasoning capabilities across diverse domains.
    \item Our fine-tuned model, \textsc{ProofSeek}, outperforms DeepSeek on an unseen problem domain, achieving a 3\% improvement in proof success rate while demonstrating a 20\% reduction in execution time
    \item We successfully apply our framework to verify the correctness of AWS S3 bucket policies, showcasing its potential for automated theorem proving in practical applications.

\end{itemize}
% \end{spacing}
 \vspace{-0.5em}
\section{Background and Related Work}
%{\color{blue} ProofAug 

%Hallucinations, LLMs and use cases.(need for a tool like this)}

\paragraph{Formal mathematics and verification}
%Description of Isabelle/Isar Proof Language {\color{blue} Use ProofAug}
%The go into the proof environment  {\color{blue} Use Poetry}
Formal mathematics is the practice of expressing mathematical statements, proofs, and reasoning in a rigorous language that can be verified for correctness by a computer~\citep{polu2022formal}. This approach is fundamental to formal verification, which ensures the accuracy of both mathematical proofs and complex systems, including software and hardware~\citep{avigad2010understanding}. Interactive theorem provers (ITPs) such as Isabelle, Coq, Lean, and HOL Light assist with formalizations by allowing users to encode proofs in a formal language and automatically verify their correctness. 

\paragraph{Autoformalization}
Autoformalization is the process of automatically translating from natural language
statements and mathematics to formal specifications and proofs and has gained significant attention in recent years~\citep{wu2022autoformalization}.
Progress in autoformalization systems accelerate the development of tools for mathematical reasoning which can be used in machine learning without the need for
associated ground-truths~\citep{wu2022autoformalization,polu2022formal}. Large language models (LLMs) have shown promising results in this area, demonstrating the ability to translate mathematical competition problems into formal specifications in systems like Isabelle/HOL~\citep{jiang2022draft,wang2023lego,liu2025efficient}. There are strong argument that autoformalization is a promising path for systems to learn sophisticated, general purpose reasoning in all domains of mathematics and computer science~\citep{szegedy2020promising}.

\paragraph {Automated theorem proving for proof assistants}
%{\color{blue} A unified view of theorem proving with generative language models - ProofAug, FVEL}
% The community working in the field of automated theorem proving has built formal languages to describe mathematical statements, problems, and theorems and to express their respective solution processes, allowing deductive reasoning to achieve the final answers or proofs with rigorous verification. 
Tools like Sledgehammer~\citep{bohme2010sledgehammer} automate reasoning within the interactive theorem prover Isabelle by translating goals into other types of logic, which are then sent to automated theorem provers like Z3~\citep{de2008z3} and Vampire~\citep{riazanov2001vampire}. If they find a proof, Sledgehammer reconstructs it in an applicable format~\citep{zhao2024subgoalxl}. PISA (Portal to ISAbelle)~\citep{jiang2021lisa} supports automated proof search for Isabelle and can be used to run multiple instances of Isabelle for concurrent checking. 
% Such tools however 
% While these tools have significantly advanced automated reasoning, 
% rely heavily on symbolic methods and predefined heuristics, limiting their flexibility in generating novel proof strategies for generalized uses beyond strictly mathematical applications.
% Additionally, effectively using these tools presents a high barrier to entry in terms of technical and mathematical expertise. To address these limitations,

\paragraph{Machine learning for automated theorem proving} Recent efforts have integrated large language models with theorem proving~\citep{jiang2021lisa, jiang2022thor, wang2023lego}.
% These approaches use a combination of symbolic and sub-symbolic systems with creative prompt engineering techniques to generate verifiable proofs. By leveraging the advanced language capabilities of LLMs, these methods show promise for formalized applications such as formal verification~\citep{lin2024fvel}.
Such techniques build capable LLMs as black-box distribution generators that suggest proof steps or whole proofs, which are then verified by an interactive proof system. The first step is to build a paradigm that captures the task of theorem proving in the context of language modeling formulated as a triple \( (\mathcal{A}, S, T) \) where $A \subset \Sigma^*$ is the set of proof steps, $S$ is the set of proof states, $T : S \times A \to S$ is the state transition function which applies proof steps to states~\citep{liu2025efficient}.
% \vspace{1pt} \begin{spacing}{0.8} 
% \begin{itemize}
% % \setlength{\itemsep}{0pt}
% % \setlength{\parskip}{0pt}  
% % \setlength{\topsep}{0pt} 
%     \item $A \subset \Sigma^*$: The set of proof steps (i.e., logical deductions, transformations).
%     \item $S$: The set of states, where each state represents the state of the proof (including metadata such as success, failure, or errors and so on).
%     \item $T : S \times A \to S$: The state transition function, which applies proof steps to states.
% \end{itemize}
% \end{spacing}
To use LLMs as black boxes or capable proof assistants~\citep{agrawal2022towards}, when provided with a theorem statement $x_f \in \Sigma^*$, it needs to provide a valid proof $y_f \in \Sigma^*$ which is valid if applying the proof steps results in a terminal state where $s_{x_f} \parallel y_f .\text{finish} = \text{True}$.

% \renewcommand{\arraystretch}{0.9} 
% \begin{table}[t]
% \caption{Summary of existing neural theorem proving approaches}
%    \centering
%    \small
%    \renewcommand{\arraystretch}{1.2} % Adjust row spacing
%   \begin{tabularx}{\textwidth}{p{3cm} p{6cm} p{5cm}}
%        \toprule
%        \textbf{Generation Method}  & \textbf{Prompt Components} & \textbf{Expected Output}  \\
%        \midrule
%        \multicolumn{3}{l}{\textbf{Single-Pass Methods}} \\
%        Naive  & $x_f$ & Full proof  \\
%        DSP  & $x_f$ + informal draft & Proof sketch \\
%        LEGO  & $x_f$ + draft + skills & Skill completion + proof sketch  \\
%        Lyra  & $x_f$ + last attempt error & Proof sketch  \\
%        \midrule
%       \multicolumn{3}{l}{\textbf{Proof-Step Methods}} \\
%        GPT-f  & Selected goal & Tactic  \\
%        ReProver  & Retrieved premises + ps & Tactic  \\
%        POETRY  & Context + proof state (w/ ‘sorry’) & Proof step  \\
%        ProofAug  & $x_f$ + informal draft & Full proof  \\
%        ProofAug + ERP & Context + proof state & Proof of conjecture \\
%        \bottomrule
%    \end{tabularx} %   \caption{Comparison of LLM based proof generation methods.~\cite{liu2025efficient}}
%    \label{tab:proof_methods}
% \end{table}

\paragraph{Neural theorem proving}
%An In-Context Learning Agent for Formal Theorem-Proving
Neural theorem provers combine neural language models (LLMs) with symbolic proof assistants to address formal mathematical tasks. Early implementations focused on premise selection~\citep{irving2016deepmath} which have been shown to be highly effective in guiding proof searches~\citep{wang2017premise}. Proof search strategies~\citep{polu2020generative} explore the space of possible proofs by generating intermediate steps or tactics. 
% For instance, the theorem prover E~\cite{loos2017deep} has achieved strong results on Mizar statements by employing efficient proof search methods. Similarly, Sledgehammer~\cite{paulson2023sledgehammer}, a widely used tool in Isabelle, integrates proof search with external automated provers to improve proof discovery. Recent advances in neural theorem proving focus on integrating the generative capabilities of LLMs with structured proof search algorithms, enabling models to generate proofs step by step or even in a single pass.
Two primary methodologies for neural theorem proving have emerged: single-pass methods and proof-step methods. Single-pass methods such as DSP~\citep{jiang2022draft}, LEGO-PROVER~\citep{wang2023lego}, and Lyra~\citep{zheng2023lyra} aim to generate entire proofs at once using prompts enriched with contextual information.
% \vspace{1pt} \begin{spacing}{0.8} 
% \begin{itemize}
% % \setlength{\itemsep}{0pt}
% % \setlength{\parskip}{0pt}  
% % \setlength{\topsep}{0pt} 
%     \item Naive: Generates full proofs directly from formal theorem statements ($x_f$) and informal drafts.
%     \item DSP: Produces proof sketches based on $x_f$ and informal drafts~\cite{jiang2022draft}.
%     \item LEGO-PROVER: Incorporates modular skills and error correction mechanisms to complete proofs incrementally~\cite{wang2023lego}.
%     \item Lyra: Combines $x_f$ with feedback from previous errors to refine proof sketches~\cite{zheng2023lyra}.
% \end{itemize}
% \end{spacing}
Proof-step methods decompose the proving process into incremental steps. These methods, such as GPT-f~\citep{polu2020generative} and POETRY~\citep{wang2025proving} utilize LLMs to generate individual tactics or proof steps conditioned on the current state of the proof environment.
% \vspace{1pt} \begin{spacing}{0.8} 
% \begin{itemize}
% % \setlength{\itemsep}{0pt}
% % \setlength{\parskip}{0pt}  
% % \setlength{\topsep}{0pt} 
%     \item GPT-f: Selects goals and generates corresponding tactics~\cite{polu2020generative}.
%     \item POETRY: Refines individual steps using placeholders like "sorry" to handle incomplete proofs~\cite{wang2025proving}.
% \end{itemize}
% \end{spacing}

\paragraph{Reinforcement learning for theorem proving}
Reinforcement Learning (RL) for theorem proving emphasizes model learning through direct feedback via trial and error. Early approaches utilizing RL had been unsuccessful due to the infinite action space as well as the absence of a direct self-play setup~\citep{polu2022formal}.
% Until recently, the naive application of reinforcement learning to formal mathematics had been unsuccessful due to the infinite action space as well as the absence of a direct self-play setup~\cite{polu2022formal}. 
Recently RL approaches have been met with success with the introduction of powerful training regimes like Direct Preference Optimization (DPO)~\citep{rafailov2023direct}, Proximal Policy Optimization (PPO)~\citep{schulman2017proximal}, and GRPO~\citep{shao2024deepseekmath}.
% mathematical reasoning, and, by extension, theorem proving in LLMs has improved significantly.
RL frameworks for theorem proving model interactions between LLMs and generated proofs as a Markov Decision Process (MDP). The reward function is typically designed around binary proof completion: assigning a reward of 1 if the proof is verified as correct and 0 otherwise~\citep{dong2024formal}. This binary reward system provides clear feedback for optimizing performance. Other approaches incorporate search algorithms into RL frameworks, such as Best-First Search (BFS) which guides proof generation by prioritizing promising paths based on heuristic evaluations~\citep{yang2023leandojo}; and Monte Carlo Tree Search (MCTS) which explores potential proof paths systematically by balancing exploration and exploitation~\citep{lample2022hypertree}.
% % \vspace{1pt} \begin{spacing}{0.8} 
% \begin{itemize}
% % \setlength{\itemsep}{0pt}
% % \setlength{\parskip}{0pt}  
% % \setlength{\topsep}{0pt} 
%     \item Best-First Search (BFS): Guides proof generation by prioritizing promising paths based on heuristic evaluations~\cite{yang2023leandojo}.
%     \item Monte Carlo Tree Search (MCTS): Explores potential proof paths more systematically by balancing exploration and exploitation~\cite{lample2022hypertree}.
% \end{itemize}
% % \end{spacing}
RL-based approaches aim to improve both single-pass and stepwise proof generation models. In single-pass methods, RL optimizes full-proof generation by rewarding logical consistency using Chain-of-Thought tokens and verification success.
In stepwise methods, RL enhances tactic prediction by refining intermediate steps based on feedback from symbolic verifiers. Recent works such as DeepSeekMath~\citep{xin2024deepseek} demonstrate that reinforcement learning can significantly enhance models' reasoning abilities by improving their capacity to generate coherent and verifiable proofs over time. %Table \ref{tab:proof_methods} gives an overview of proof generation methods using LLMs that are relevant.

\section{Method}
\begin{figure}[t]
    \centering
    \begin{minipage}{0.496\textwidth} 
        \centering
        \includegraphics[width=\linewidth]{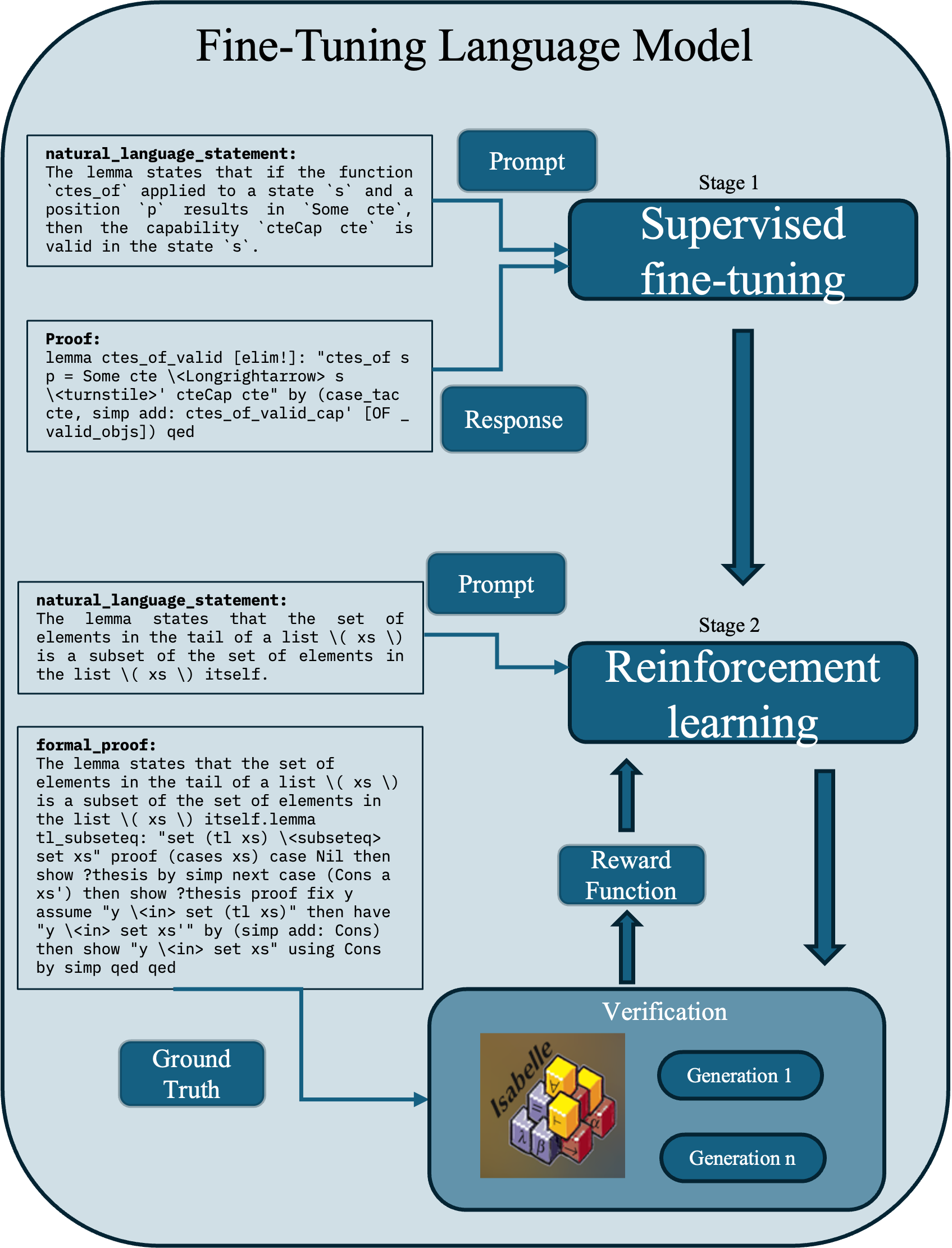}
        %\caption{}
        % \label{fig:Finetunig}
    \end{minipage}
    \hfill
    \begin{minipage}{0.496\textwidth}
        \centering
        \includegraphics[width=\linewidth]{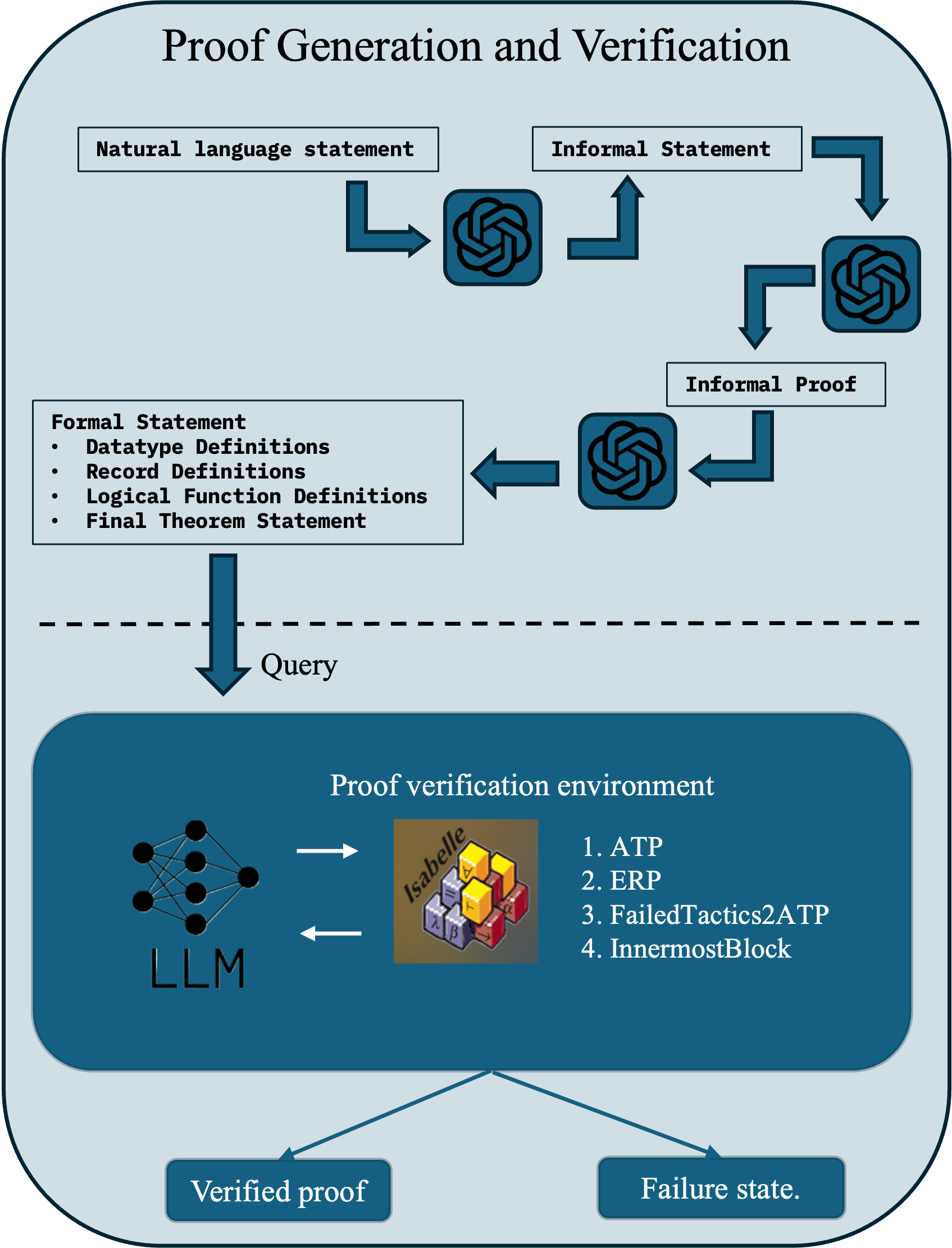}
        %\caption{Proof Generation and Verification}
        % \label{fig:Generation}
    \end{minipage}
    \caption{The two core components within the \textsc{ProofSeek} framework: (a) the fine-tuning language model module, (b) the proof generation and verification module}
    \label{fig:approach}
\end{figure}

Building on the work of the workflow described in DSP~\citep{jiang2022draft} and the proof construction method in ProofAug~\citep{liu2025efficient}, in this section we present our framework \textsc{ProofSeek} that leverages the strengths of proof-step and whole-proof generation, as well as the natural language generation paradigms of LLMs. 
\textsc{ProofSeek} consists of two core components: a component for fine-tuning a language model using SFT and RL, and a proof generation and verification component for generating and building the formal proofs.  Both components of \textsc{ProofSeek} are shown in Figure~\ref{fig:approach}.
% leverage SFT to make the model better at syntactic generation, then we leverage reinforcement learning to make it better at semantic generation.
\textbf{Our framework is generalizable across domains where the input is a mathematical statement, policy code, or natural language statement, and the output is a verified proof state or a failure state}. The details of  \textsc{ProofSeek} is shown in Algorithm~\ref{alg:proof_verification}. We first fine-tune a whole-proof generation model using our two-stage approach. Then, we build a formal statement that represents the provided policy code or mathematical statement. Finally, we leverage the fine-grained proof structure analysis method of ProofAug to verify the generated formal statements. 

\subsection{Fine-Tuning Language Model}
We first discuss how we fine-tune both the supervised and reinforcement learning stages of our approach (Figure~\ref{fig:approach}(a), Algorithm~\ref{alg:proof_verification} lines 1-2). For our training task, we opt to fine-tune our model for whole-proof generation, treating the construction of formal proofs as a general code completion task. The model aims to generate entire proof code based on a theorem in a single step. This scalable approach has been shown to be effective for both training and inference deployment~\citep{xin2024deepseek}.
Several recent efforts have explored training models specifically for theorem proving, including LLEMMA, which was pretrained using Code Llama on Proof-Pile-2 and outperformed other open-source models on benchmarks such as MATH and Minerva~\citep{lewkowycz2022solving}. Given the availability of more capable open-source models, we select DeepSeek-Math-7B-rl~\citep{shao2024deepseekmath} as our base model due to its superior performance in mathematical reasoning tasks.
Our two-stage fine-tuning process consists of Supervised Fine-Tuning (SFT) on curated theorem-proof pairs, followed by Reinforcement Learning (RL) Fine-Tuning using correctness-based reward signals. By integrating formal proof checking into the reward mechanism, we enhance the model’s ability to generate valid and verifiable proofs.

\subsubsection{Dataset construction}
For the two separate stages of fine-tuning, we build different datasets for each purpose. To ensure there are no data leaks during training (i.e., the RL training already sees a particular statement-proof pair), we filter the dataset using a PISA setup. We start with FVEL\textsubscript{\textnormal{ER}}, which includes verification tasks formulated in Isabelle, containing 758 theories, 29,125 lemmas, and 200,646 proof steps in total, with in-depth dependencies~\citep{lin2024fvel}. This translates to 26,192 statements accompanied by their proofs and proof steps. As noted by~\cite{lin2024fvel}, the proofs contain complex dependencies. Due to the setup of the RL training environment, the dataset we need requires ``self-contained" proofs or complete proofs that return a verified proof state without further processing. To accommodate this, we filter FVEL\textsubscript{\textnormal{ER}} using PISA to retain only the proofs that return true when iterated through the dataset. The other proofs, while correct, require additional dependencies. The filtered dataset, containing 1,138 statement-proof pairs, was used for the RL stage.

Meanwhile, the remaining 25,054 pairs were allocated for the supervised fine-tuning stage. Due to the nature of our intended application---generalizability for proof generation across domains---we needed to curate a dataset for instruction tuning that included a natural language description of the statement to be proved and its corresponding proofs. We processed this dataset using OpenAI API calls (GPT-4o) to build an SFT dataset in the form (\textit{proof}, \textit{statement}, \textit{natural\_language\_statement}). For this iteration, we selected 2,000 samples from the 25,054 unproven statements for API prompting. After this process, we arrived at 1,981 samples for the SFT stage and, in a similar fashion, 1,138 samples in the form (\textit{natural\_language\_statement}, \textit{formal\_proof}) for the RL stage.

\subsubsection{Supervised fine-tuning}

We first fine-tune a large language model to generate formal proofs in the Isabelle theorem prover. We leverage Unsloth's optimized training framework for parameter-efficient fine-tuning (PEFT). The model is trained using instruction tuning, where each training instance consists of theorem statements paired with corresponding Isabelle proofs to ensure structured learning. We use the FastLanguageModel library from Unsloth, which supports high-efficiency fine-tuning and inference. Instead of fine-tuning the full model, we use Low-Rank Adaptation (LoRA)~\citep{hu2022lora} to fine-tune the relevant portions, significantly reducing computational overhead while maintaining high performance. We initialize DeepSeek-Math-7B-rl as the base model. We fine-tune our model using SFTTrainer. After training, the fine-tuned model is published to the Hugging Face Model Hub for further tuning.

\subsubsection{Reinforcement learning}

In the second stage of our fine-tuning approach, we employ Group Relative Policy Optimization (GRPO)~\citep{shao2024deepseekmath} as our reinforcement learning (RL) algorithm, which has demonstrated superior effectiveness and efficiency compared to Proximal Policy Optimization (PPO)~\citep{schulman2017proximal}. Unlike PPO, which updates a policy based on absolute reward values, GRPO optimizes model updates by considering the relative ranking of multiple generated outputs, making it well-suited for structured text generation tasks.

GRPO operates by sampling multiple candidate proofs for each theorem prompt and optimizing the model based on relative rewards assigned to outputs within the group. This method improves training stability and encourages the generation of higher-quality proofs by leveraging pairwise ranking rather than relying solely on absolute correctness metrics. The reward function is designed to interact with PISA to verify the generated proofs.

We implement GRPO using Unsloth's FastLanguageModel (similar to the SFT stage), for high-efficiency training with parameter-efficient fine-tuning (PEFT) using Low-Rank Adaptation (LoRA). To enhance proof validity and structure, we employ two reward functions:
% \vspace{1pt} \begin{spacing}{0.8} 
\begin{enumerate}
    \item Correctness Reward: Extracts the Isabelle proof from the model's response and compares it to the ground truth proof.
    \item Formal Proof Verification via PISA: Utilizes PISA for proof checking. If a generated proof verifies in Isabelle, it receives a reward of 1; otherwise, 0.
\end{enumerate}
% \end{spacing}
Finally, the trained model is uploaded to the Hugging Face Model Hub for inference\footnote{The model can be downloaded from: \url{https://huggingface.co/kings-crown/ProofSeek_v1}}.

\begin{algorithm}[t]
\caption{\textsc{ProofSeek}}\label{alg:proof_verification}
\begin{algorithmic}[1]
\Require Natural language statement $s$, fine-tuned whole-proof generation model $\pi_{\theta}$, proof verification environment.
% \Statex
\Statex \textbf{Fine-Tune Model:} Train a whole-proof generation model $\pi_{\theta}$ using a two-stage approach.
\State \quad $\pi_{\theta} = \arg\max_{\theta} \mathbb{E}_{(x, y) \sim \mathcal{D}} [\log P_{\theta} (y | x)]$
\State \quad where $\mathcal{D}$ is the dataset containing natural language statements and their proofs.
% \Statex
\Statex \textbf{Autoformalization:} Use a structured workflow to construct a formal statement $S$ from the provided code or mathematical statement by generating:
\State \quad Datatype definitions $\mathcal{D} = \{D_i\}$ capturing entities involved.
\State \quad Record definitions $R = \{r_j\}$ for structured objects (e.g., access policies).
\State \quad Logical function definitions $F = \{f_k: D_i \to D_j\}$.
\State \quad The final theorem statement as $\forall x \in X, P(x) \Rightarrow Q(x)$.
% \Statex
\Statex \textbf{Proof Construction:} Apply ProofAug for verification
\State \quad Model proof generation as a state transition system $(\mathcal{A}, S, T)$:
\State \quad \quad $\mathcal{A}$: Finite set of proof actions (inference steps).
\State \quad \quad $S$: Finite set of proof states.
\State \quad \quad $T: S \times \mathcal{A} \to S$: Transition function under inference rules.
\State \quad Generate proof steps iteratively using $\pi_{\theta}$
\State \quad \quad \( a^* = \arg\max_{a \in \mathcal{A}} P(a | S) \), during  inference
\State \quad Validate each step \( a[i] \) using ATP:
\State \quad \quad \( \textit{error} \leftarrow T(s_{\text{this}}, \langle ATP \rangle). \textit{error} \)
\State \quad If ATP fails, apply ERP correction:
\State \quad \quad \( y_f' \sim \pi(p(x_i || y_i, x_f || y_f)) \)
\State \quad If ERP fails, apply heuristic tactics:
\State \quad \quad \( y_f' \leftarrow \textsc{FailedTactics2ATP}(y_f) \)
\State \quad If no valid step exists, backtrack to the last valid proof block:
\State \quad \quad \( \textit{block} \leftarrow \textsc{InnermostBlock}(i, a) \)
\State \quad Terminate when a valid proof state or failure state $\emptyset$ is reached.
\State \Return Verified proof or failure state.
\end{algorithmic}
\end{algorithm}

\subsection{Autoformalization}

The first stage of verification in \textsc{ProofSeek} involves the autoformalization content being verified (Algorithm~\ref{alg:proof_verification} lines 3-6). We employ semantic parsing to translate a natural language statement into its logical form. We do this by generating intermediary stages, turning the natural language statement into an informal statement, then deriving an informal proof and a formal statement that is a representation of the initial statement in its logical form for the prover.

\subsubsection{Informal Representation}
We start with an informal dataset where $\mathcal{N} = {(s_i^{\mathcal{N}})}_{i=1}^{|\mathcal{N}|}$ represents natural language statements. For each $s_i^{\mathcal{N}}$, we prompt an LLM (we use GPT-4o) to generate an informal description $d_i$ to provide additional interpretability and structure for both the user and the model in the next steps. Given $(s_i^{\mathcal{N}}, d_i)$, the LLM then produces an informal proof $p_i$ in the same format through curated prompts. This informal proof $p_i$ is seen as the skeleton for the formal representation.

\subsubsection{Formal Proof Representation}

To construct a valid Isabelle/HOL proof, the logical form must accurately reflect the original statement. The transformation from natural language to formal representation follows a structured pipeline, leveraging incremental representation generations (which~\cite{jiang2022draft} showed to be more efficient than single-shot representation generation). The generated informal proof $p_i$ allows the generated formalization $S$ to be more consistent and faithful. Our approach ensures that the generated proof includes:
% \vspace{1pt} \begin{spacing}{0.8} 
\begin{itemize}

    \item \textbf{Datatype Definitions} ($\mathcal{D} = \{D_i\}$): Define structured entities present in the statement. These serve as the foundational building blocks for formal reasoning.

    \item \textbf{Record Definitions} ($R = \{r_j\}$): Represent structured objects, such as access policies or logical relations. Used to define attributes and relationships between entities.
    
    \item \textbf{Logical Function Definitions} ($F = \{f_k: D_i \to D_j\}$): Encode logical operations and transformations essential for proof construction. Define predicates and functions that express constraints and properties.

    \item \textbf{Final Theorem Statement} ($\forall x \in X, P(x) \Rightarrow Q(x)$): The theorem statement that encapsulates the key property to be proved. This formally expresses the intended logical relationship in a structured way.
\end{itemize}
% \end{spacing}
We employ stepwise prompting to sequentially construct the theorem statement:
\begin{enumerate}
\item \textbf{Natural Language Input} ($s$): The initial informal statement describing what should be proved.
\item \textbf{Informal Description} ($d$): A structured interpretation that clarifies the semantics of $s$.
\item \textbf{Informal Proof} ($p$): High-level reasoning outlining requirements to guide formalization.
\item \textbf{Formal Statement} ($S$): A logically rigorous theorem statement, translation of $p$ into theorem-prover-compatible syntax.
\end{enumerate}

%\begin{enumerate}
%\item Datatype definitions: Definitions capturing the structure of the entities involved in the statement.
%\item Record definitions: Representations of structured objects related to access policies.
%\item Function definitions: Logical functions needed to express the property being proved.
%\item Final theorem statement: A theorem that states a key property derived from the given access policy.
%\end{enumerate}

%The prompt used to generate this stage needs to be configured to the nature of the statements being verified. Through incremental prompting, we generate, given a natural language statement $s$, its description $d$, the informal proof $p$, and its formal statement $S$.

\subsection{Proof Construction}

Once the formal statement \( S \) is generated, we employ an interactive theorem prover (ITP)-based approach, using prompting inspired by~\cite{jiang2022draft}, to guide proof-step generation and construction (Algorithm~\ref{alg:proof_verification} lines 7-22). We use ProofAug~\citep{ liu2025efficient}, an augmentation strategy that refines proofs by integrating automated theorem proving (ATP), efficient recursive proving (ERP), and heuristic-based tactic generation. The proof construction process in ProofAug that we use follows a structured pipeline, which we summarize as follows:%{\color{red} This is essentially the breakdown of algorithm 2 in the ProofAug paper}
% \vspace{1pt} \begin{spacing}{0.8} 
\begin{enumerate}
    \item \textbf{Proof representation in ITP:} Model the proof as a state transition system  \( (\mathcal{A}, S, T) \).
    \item \textbf{Proof-step generation via language models:}  A fine-tuned generative language model \( \pi_{\theta} \) predicts valid proof steps \( a \in \mathcal{A} \) conditioned on the proof state:
    \[
    \pi_{\theta}(x) = \arg\max_{a \in \mathcal{A}} P(a | S)
    \]
     \item \textbf{ATP validation:} Each generated proof step is validated using ATPs. If a proof step is not trivially correct and has \textit{sorry} proofs the system invokes an ATP-based evaluation:
    \[
    \textit{error} \leftarrow T(s_{\text{this}}, \langle ATP \rangle). \textit{error}
    \]  
    \item \textbf{Efficient recursive proving via ERP module:} If ATP validation fails, ERP attempts an alternative inference:
    \[
    y_f' \sim \pi(p(x_i || y_i, x_f || y_f))
    \]
    where \( y_f' \) is a corrected proof step. If ERP succeeds, the proof step is updated accordingly.
    \item \textbf{Heuristic-based tactics for failed proofs:}  
    When both ATP and ERP modules fail, heuristic-based fallback strategies are applied via:
    \[
    y_f' \leftarrow \textsc{FailedTactics2ATP}(y_f)
    \]
    attempting to construct missing proof steps via structured heuristics.
    \item \textbf{Backtracking and Proof Reorganization:}  
    If no valid proof step can be generated, the algorithm identifies the innermost valid proof block, resets proof state, and re-attempts construction:
    \[
    \textit{block} \leftarrow \textsc{InnermostBlock}(i, a)
    \]
\end{enumerate}
% \end{spacing}
Given a theorem statement \( x_f \), the system iteratively applies proof steps \( a \in \mathcal{A} \) until the proof reaches a valid terminal state, yielding either: a \textbf{successful proof} (\( P \)), where all proof steps are verified; or a \textbf{failure state} (\( \emptyset \)), if the proof cannot be completed.

%Once the formal statement $S$ is generated, we leverage an ITP-based approach inspired by~\cite{jiang2022draft, liu2025efficient} to guide proof-step generation and proof construction. This involves:
%\begin{itemize}
    %\item Constructing a proof object in an interactive theorem prover (ITP), modeled as a state transition system $(\mathcal{A}, S, T)$ where:
          %\begin{itemize}
              %\item $\mathcal{A}$ is the set of proof steps.
              %\item $S$ is the set of proof states.
              %\item $T : S \times \mathcal{A} \to S$ is the state transition function.
          %\end{itemize}
    %\item Using a generative language model $\pi(x)$ that produces valid proof steps given a proof state.
    %\item Employing probabilistic modeling and gradient estimation techniques to refine proof generation~\cite{liu2025efficient}.
%\end{itemize}

%Given a theorem statement $x$, the system iteratively applies proof steps $a \in \mathcal{A}$ until the proof reaches a valid terminal state, either a successful proof or failed termination.

\section{Experiments}
To evaluate the utility of our \textsc{ProofSeek} framework and assess any enhancements to the model's theorem proving capability, we aim to answer the following research questions: \\
\noindent\textbf{RQ1:} How effective is the \textsc{ProofSeek} framework at autoformalization and generating proofs in an unseen problem domain? \\
\noindent\textbf{RQ2:} How effective is the fine-tuning approach within \textsc{ProofSeek} in enhancing the theorem proving capabilities of a language model?
% \begin{itemize}
%     \item How effective is our framework at autoformalization and generating proofs in an unseen problem domain?
%     \item How effective is our finetuning approach to enhance theorem proving capabilities of a language model?
% \end{itemize}

\subsection{Experiment Setup}
To fine-tune our models and evaluate our framework, we utilize two machines (for running concurrent processes). We set up a PISA environment~\citep{jiang2021lisa} to interact with Isabelle 2022. Our $\langle ATP \rangle$ method uses 8 Isabelle proof methods (auto, simp, auto,
blast, fastforce, eval, sos, arith, simp:field simps, simp add:mod simps) as well as Sledgehammer. For the verification process, we use 4 instances of PISA and similar to~\cite{jiang2022draft} we set set the timeout for any proof step and Sledgehammer as 10 seconds and 40 seconds, respectively. We run our experiments on: AMD EPYC 7763 64-Core Processor CPU @ 2.49GHz with a NVIDIA A40-48Q and an AMD Ryzen Threadripper PRO 5975WX 32-Core Processor CPU @ 7.00GHz with 2 NVIDIA RTX A6000.

To evaluate our framework, we use two language models for comparison: (1) Deepseek-Math-7b-base (as the base-case) to reproduce the results of ProofAug~\citep{liu2025efficient}; and (2) our fine-tuned model (fine-tuned using the \textsc{ProofSeek} framework) to evaluate improvements we can achieve using their method. We follow an approach similar to~\cite{liu2025efficient} and opt to use 1-shot prompting. To keep the evaluation consistent, we limit the sample budget across all the tests to 10, and consistently use a sampling temperature $T = 0.6$ with top $p = 0.95$.

\subsubsection{Benchmark on miniF2F-test}
We evaluate our approach using the miniF2F-Test dataset\citep{zheng2021minif2f}, which includes
488 formal mathematical problems which encompasses high-school level exercises
and competition problems. The dataset is split into a validation set and a test set, each containing 244 problems~\citep{xin2024deepseek}. We
use the Isabelle part of the miniF2F-test dataset that contains an additional informal statement and informal draft for each problem~\citep{jiang2022draft}.

\subsubsection{Case Study: Verifying Correctness of AWS S3 Bucket Policies}

Amazon Web Services (AWS) allows users to create access control policies for managing access to AWS services. These policies regulate access through declarative statements that specify whether a given access control request should be allowed or denied.
% A \textit{statement} consists of a 5-tuple (\textit{Principal, Effect, Action, Resource, Condition}) where
% \noindent 
% \begin{itemize}
%     \item \textit{Principal} specifies a list of users, entities, or services
%     \item \textit{Effect} $\in$ \{Allow, Deny\} specifies whether the statement allows or denies access
%     \item \textit{Action} specifies a list of actions
%     \item \textit{Resource} specifies a list of resources
%     \item \textit{Condition} is an optional list of conditions that further constrain how access is allowed or denied.
% \end{itemize}
Given an access control request and an associated policy, access is granted \emph{if and only if} there exists at least one statement in the policy that allows the access and no statement that explicitly denies it. Thus, to verify the correctness of a policy, one must reason about the logic of the policy statements and determine whether the intended behavior (as expressed in natural language) matches the formal semantics of the policy. If the intent aligns with the semantics, the policy is considered correct with respect to that intention. Otherwise it is incorrect. For full details of the AWS policy language, we refer the reader to~\cite{BBC18}.

In this case study, we consider a set of S3 bucket policies along with a set of generated natural language description of their intended behaviors. We evaluate the effectiveness of our approach by generating Isabelle proofs that demonstrate the correctness of the policies with respect to their intended access control intents. We consider two scenarios:\\
\noindent\textbf{Scenario 1: Evaluation with manually curated dataset}
For the first phase of our analysis, we investigate the proving ability of our setup using a small, manually curated dataset of S3 bucket access policies. We randomly pick 25 policies from the Quacky dataset~\citep{eiers2022quacky} and construct formal statements for each policy code in the proof environment. We extract key components from the policy (Actions, Resources, Effects, Conditions) and convert them into formal logic types. Then, we finally transform the policy components into formal entries for Isabelle. The complete evaluation comprises 25 policy-formal statement pairs. \\
\noindent\textbf{Scenario 2: Evaluation on LLM generated dataset}
To evaluate the utility of the framework as a whole, we automate the formalization process of a set of AWS S3 Bucket Policies in a csv file using GPT-4o to generate formal statements to be proved. We use an approach inspired by DSP~\citep{jiang2022draft} to iteratively prompt the LLM to autoformalize the policy code as described in our framework. We use some of the manually curated policy-formal statement pairs as few-shot examples to guide the model along. The final dataset comprises of 243 policies, and we use the method described in ProofAug with our model to prove the statements.

\subsection{Experiment Results}

Here we present the results of our experiments on different datasets, including the miniF2F-Test dataset for mathematical theorem proving and the Quacky dataset for AWS S3 bucket policy verification\footnote{The generated proof construction jsonl files are attached in the code repository}. Our evaluation primarily focuses on the proof success rate, the average number of attempts required, and the total execution time.

Table~\ref{tb1} reports the results on the miniF2F-Test dataset, comparing our approach, \textsc{ProofSeek}, against DeepSeek. We evaluate both models with and without efficient recursive
proving (ERP). \textsc{ProofSeek} achieves a 40.1\% success rate without ERP, slightly lower than DeepSeek (41.8\%). The total execution time for ProofSeek (ERP: 38,651.78s, No ERP: 36,557.68s) is slightly lower than DeepSeek with ERP (53,658.5s).

For manually curated AWS S3 bucket policies from the Quacky dataset, \textsc{ProofSeek} achieves a 96.0\% success rate across all settings. This is unsurprising as the proof formulations were of high quality making it easy for the prover. This is also reflected in the times: \textsc{ProofSeek} (No ERP) is the fastest, completing in 196.01s, compared to DeepSeek (ERP: 573.84s, No ERP: 375.67s), demonstrating superior efficiency. When evaluating \textsc{ProofSeek} on LLM-generated policy statements: \textsc{ProofSeek} (No ERP) outperforms all other settings with a 69.1\% success rate, surpassing DeepSeek.

\begin{table}[t]
    \centering
    \caption{Experimental Results on Evaluation Datasets}\label{tb1}
    \resizebox{\textwidth}{!}{ 
    \begin{tabular}{l c c c}
        \toprule
        \textbf{Method} & \textbf{Success Rate (\%)} & \textbf{Avg Attempts} & \textbf{Total Exec. Time (h:mm:ss)} \\
        \midrule
        \multicolumn{4}{l}{\textbf{MiniF2F Dataset (245 Problems)}} \\
        \textsc{ProofSeek} (Ours) (ERP)      & 41.8 & 5.66 & 10:44:11 \\
        \textsc{ProofSeek} (Ours) (No ERP)   & 40.1 & 5.67 & 10:09:17  \\
        DeepSeek (ERP)                       & 42.2 & 5.59 & 14:54:18  \\
        DeepSeek (No ERP)                    & 41.8 & 5.54 & 07:24:12  \\
        \midrule
        \multicolumn{4}{l}{\textbf{Curated Quacky Dataset (25 Problems)}} \\
        \textsc{ProofSeek} (Ours) (ERP)      & 96.0 & 0.60 & 00:06:33  \\
        \textsc{ProofSeek} (Ours) (No ERP)   & 96.0 & 0.44 & 00:03:16  \\
        DeepSeek (ERP)                       & 96.0 & 0.64 & 00:09:33  \\
        DeepSeek (No ERP)                    & 96.0 & 0.84 & 00:06:15  \\
        \midrule
        \multicolumn{4}{l}{\textbf{Generated Quacky Dataset (243 Problems)}} \\
        \textsc{ProofSeek} (Ours) (ERP)      & 66.6 & 1.33 & 00:20:34  \\
        \textsc{ProofSeek} (Ours) (No ERP)   & \textbf{69.1} & 1.15 & \textbf{00:20:27} \\
        DeepSeek (ERP)                       & 63.3 & 2.05 & 00:26:52  \\
        DeepSeek (No ERP)                    & 66.6 & 1.95 & 00:24:36  \\
        \bottomrule
    \end{tabular}}
\end{table}

\noindent\textbf{MiniF2F-Test Performance (RQ1)}: Our framework demonstrates effectiveness in autoformalization and proof generation in unseen domains, achieving performance comparable to DeepSeek while improving computational efficiency. \\
\noindent\textbf{AWS S3 Policy Verification (RQ2)}: Our system’s fine-tuning results in highly efficient and accurate verification of structured policies, confirming enhancements in theorem proving capabilities. Moreover, \textsc{ProofSeek} proves more effective on LLM-generated formalizations, demonstrating robustness in handling formalization related tasks in generalized problem domains and maintaining higher success rates over DeepSeek.

\section{Discussion and Future Work}
Although our framework demonstrated its practical utility for real-world use cases, our results on benchmark datasets remain behind SOTA approaches. Additionally, we speculate that the fine-tuning process enhanced the theorem-proving capability of the language model, as evidenced by the smaller number of proof attempts required to complete proofs. However, due to the construction of the symbolic methods being used, it did not perform as well on the benchmark as expected. Moreover, from our initial results, it is clear that further fine-tuning is necessary, both using supervised and RL-based methods. In this work, we refrained from further training so as to not overfit. We also look forward to incorporating reasoning-based feedback \citep{xie2025teachinglanguagemodelscritique} to build better models for formal verification.

We believe that an important direction for future work is to fully leverage the reliability aspect of language model-generated proofs across systems. Reliability is a highly desirable property in inherently probabilistic systems. In the future, we aim to incorporate other forms of symbolic systems, such as knowledge graphs, to make automated theorem proving with LLMs more reliable and consistent.

\bibliography{nesy2025}

\begin{thebibliography}{37}
\providecommand{\natexlab}[1]{#1}
\providecommand{\url}[1]{\texttt{#1}}
\expandafter\ifx\csname urlstyle\endcsname\relax
  \providecommand{\doi}[1]{doi: #1}\else
  \providecommand{\doi}{doi: \begingroup \urlstyle{rm}\Url}\fi

\bibitem[Agrawal et~al.(2022)Agrawal, Gadgil, Goyal, Narayanan, and Tadipatri]{agrawal2022towards}
Ayush Agrawal, Siddhartha Gadgil, Navin Goyal, Ashvni Narayanan, and Anand Tadipatri.
\newblock Towards a mathematics formalisation assistant using large language models.
\newblock \emph{arXiv preprint arXiv:2211.07524}, 2022.

\bibitem[Avigad(2010)]{avigad2010understanding}
Jeremy Avigad.
\newblock Understanding, formal verification, and the philosophy of mathematics.
\newblock \emph{Journal of the Indian Council of Philosophical Research}, 27:\penalty0 161--197, 2010.

\bibitem[Azerbayev et~al.(2023)Azerbayev, Schoelkopf, Paster, Santos, McAleer, Jiang, Deng, Biderman, and Welleck]{azerbayev2023llemma}
Zhangir Azerbayev, Hailey Schoelkopf, Keiran Paster, Marco~Dos Santos, Stephen McAleer, Albert~Q Jiang, Jia Deng, Stella Biderman, and Sean Welleck.
\newblock Llemma: An open language model for mathematics.
\newblock \emph{arXiv preprint arXiv:2310.10631}, 2023.

\bibitem[Backes et~al.(2018)Backes, Bolignano, Cook, Dodge, Gacek, Luckow, Rungta, Tkachu, and Varming]{BBC18}
John Backes, Pauline Bolignano, Byron Cook, Catherine Dodge, Andrew Gacek, Kasper Luckow, Neha Rungta, Oksana Tkachu, and Carsten Varming.
\newblock Semantic-based automated reasoning for aws access policies using smt.
\newblock In \emph{Proceedings of the 18th Conference on Formal Methods in Computer-Aided Design (FMCAD 2018), Austin, Texas, USA, October 30 - November 2, 2018}, pages 1--9, 2018.

\bibitem[Bibel(2013)]{bibel2013automated}
Wolfgang Bibel.
\newblock \emph{Automated theorem proving}.
\newblock Springer Science and Business Media, 2013.

\bibitem[B{\"o}hme and Nipkow(2010)]{bohme2010sledgehammer}
Sascha B{\"o}hme and Tobias Nipkow.
\newblock Sledgehammer: judgement day.
\newblock In \emph{Automated Reasoning: 5th International Joint Conference, IJCAR 2010, Edinburgh, UK, July 16-19, 2010. Proceedings 5}, pages 107--121. Springer, 2010.

\bibitem[De~Moura and Bj{\o}rner(2008)]{de2008z3}
Leonardo De~Moura and Nikolaj Bj{\o}rner.
\newblock Z3: An efficient smt solver.
\newblock In \emph{International conference on Tools and Algorithms for the Construction and Analysis of Systems}, pages 337--340. Springer, 2008.

\bibitem[Dong et~al.(2024)Dong, Mahankali, and Ma]{dong2024formal}
Kefan Dong, Arvind Mahankali, and Tengyu Ma.
\newblock Formal theorem proving by rewarding llms to decompose proofs hierarchically.
\newblock \emph{arXiv preprint arXiv:2411.01829}, 2024.

\bibitem[Eiers et~al.(2022)Eiers, Sankaran, Li, O'Mahony, Prince, and Bultan]{eiers2022quacky}
William Eiers, Ganesh Sankaran, Albert Li, Emily O'Mahony, Benjamin Prince, and Tevfik Bultan.
\newblock Quacky: Quantitative access control permissiveness analyzer.
\newblock In \emph{Proceedings of the 37th IEEE/ACM International Conference on Automated Software Engineering}, pages 1--5, 2022.

\bibitem[Guo et~al.(2025)Guo, Yang, Zhang, Song, Zhang, Xu, Zhu, Ma, Wang, Bi, et~al.]{guo2025deepseek}
Daya Guo, Dejian Yang, Haowei Zhang, Junxiao Song, Ruoyu Zhang, Runxin Xu, Qihao Zhu, Shirong Ma, Peiyi Wang, Xiao Bi, et~al.
\newblock Deepseek-r1: Incentivizing reasoning capability in llms via reinforcement learning.
\newblock \emph{arXiv preprint arXiv:2501.12948}, 2025.

\bibitem[Hu et~al.(2022)Hu, Shen, Wallis, Allen-Zhu, Li, Wang, Wang, Chen, et~al.]{hu2022lora}
Edward~J Hu, Yelong Shen, Phillip Wallis, Zeyuan Allen-Zhu, Yuanzhi Li, Shean Wang, Lu~Wang, Weizhu Chen, et~al.
\newblock Lora: Low-rank adaptation of large language models.
\newblock \emph{ICLR}, 1\penalty0 (2):\penalty0 3, 2022.

\bibitem[Irving et~al.(2016)Irving, Szegedy, Alemi, E{\'e}n, Chollet, and Urban]{irving2016deepmath}
Geoffrey Irving, Christian Szegedy, Alexander~A Alemi, Niklas E{\'e}n, Fran{\c{c}}ois Chollet, and Josef Urban.
\newblock Deepmath-deep sequence models for premise selection.
\newblock \emph{Advances in neural information processing systems}, 29, 2016.

\bibitem[Jiang et~al.(2022{\natexlab{a}})Jiang, Welleck, Zhou, Li, Liu, Jamnik, Lacroix, Wu, and Lample]{jiang2022draft}
Albert~Q Jiang, Sean Welleck, Jin~Peng Zhou, Wenda Li, Jiacheng Liu, Mateja Jamnik, Timoth{\'e}e Lacroix, Yuhuai Wu, and Guillaume Lample.
\newblock Draft, sketch, and prove: Guiding formal theorem provers with informal proofs.
\newblock \emph{arXiv preprint arXiv:2210.12283}, 2022{\natexlab{a}}.

\bibitem[Jiang et~al.(2021)Jiang, Li, Han, and Wu]{jiang2021lisa}
Albert~Qiaochu Jiang, Wenda Li, Jesse~Michael Han, and Yuhuai Wu.
\newblock Lisa: Language models of isabelle proofs.
\newblock In \emph{6th Conference on Artificial Intelligence and Theorem Proving}, pages 378--392, 2021.

\bibitem[Jiang et~al.(2022{\natexlab{b}})Jiang, Li, Tworkowski, Czechowski, Odrzyg{\'o}{\'z}d{\'z}, Mi{\l}o{\'s}, Wu, and Jamnik]{jiang2022thor}
Albert~Qiaochu Jiang, Wenda Li, Szymon Tworkowski, Konrad Czechowski, Tomasz Odrzyg{\'o}{\'z}d{\'z}, Piotr Mi{\l}o{\'s}, Yuhuai Wu, and Mateja Jamnik.
\newblock Thor: Wielding hammers to integrate language models and automated theorem provers.
\newblock \emph{Advances in Neural Information Processing Systems}, 35:\penalty0 8360--8373, 2022{\natexlab{b}}.

\bibitem[Lample et~al.(2022)Lample, Lacroix, Lachaux, Rodriguez, Hayat, Lavril, Ebner, and Martinet]{lample2022hypertree}
Guillaume Lample, Timothee Lacroix, Marie-Anne Lachaux, Aurelien Rodriguez, Amaury Hayat, Thibaut Lavril, Gabriel Ebner, and Xavier Martinet.
\newblock Hypertree proof search for neural theorem proving.
\newblock \emph{Advances in neural information processing systems}, 35:\penalty0 26337--26349, 2022.

\bibitem[Lewkowycz et~al.(2022)Lewkowycz, Andreassen, Dohan, Dyer, Michalewski, Ramasesh, Slone, Anil, Schlag, Gutman-Solo, et~al.]{lewkowycz2022solving}
Aitor Lewkowycz, Anders Andreassen, David Dohan, Ethan Dyer, Henryk Michalewski, Vinay Ramasesh, Ambrose Slone, Cem Anil, Imanol Schlag, Theo Gutman-Solo, et~al.
\newblock Solving quantitative reasoning problems with language models.
\newblock \emph{Advances in Neural Information Processing Systems}, 35:\penalty0 3843--3857, 2022.

\bibitem[Lin* et~al.(2024)Lin*, Cao*, Huang*, Wang*, Lu, Liu, Song, and Liang]{lin2024fvel}
Xiaohan Lin*, Qingxing Cao*, Yinya Huang*, Haiming Wang*, Jianqiao Lu, Zhengying Liu, Linqi Song, and Xiaodan Liang.
\newblock {FVEL}: Interactive formal verification environment with large language models via theorem proving.
\newblock In \emph{The Thirty-eight Conference on Neural Information Processing Systems Datasets and Benchmarks Track}, 2024.
\newblock URL \url{https://openreview.net/forum?id=d0gMFgrYFB}.

\bibitem[Liu et~al.(2025)Liu, Sun, Li, and Yao]{liu2025efficient}
Haoxiong Liu, Jiacheng Sun, Zhenguo Li, and Andrew~C Yao.
\newblock Efficient neural theorem proving via fine-grained proof structure analysis.
\newblock \emph{arXiv preprint arXiv:2501.18310}, 2025.

\bibitem[Loos et~al.(2017)Loos, Irving, Szegedy, and Kaliszyk]{loos2017deep}
Sarah Loos, Geoffrey Irving, Christian Szegedy, and Cezary Kaliszyk.
\newblock Deep network guided proof search.
\newblock \emph{arXiv preprint arXiv:1701.06972}, 2017.

\bibitem[Polu and Sutskever(2020)]{polu2020generative}
Stanislas Polu and Ilya Sutskever.
\newblock Generative language modeling for automated theorem proving.
\newblock \emph{arXiv preprint arXiv:2009.03393}, 2020.

\bibitem[Polu et~al.(2022)Polu, Han, Zheng, Baksys, Babuschkin, and Sutskever]{polu2022formal}
Stanislas Polu, Jesse~Michael Han, Kunhao Zheng, Mantas Baksys, Igor Babuschkin, and Ilya Sutskever.
\newblock Formal mathematics statement curriculum learning.
\newblock \emph{arXiv preprint arXiv:2202.01344}, 2022.

\bibitem[Rafailov et~al.(2023)Rafailov, Sharma, Mitchell, Manning, Ermon, and Finn]{rafailov2023direct}
Rafael Rafailov, Archit Sharma, Eric Mitchell, Christopher~D Manning, Stefano Ermon, and Chelsea Finn.
\newblock Direct preference optimization: Your language model is secretly a reward model.
\newblock \emph{Advances in Neural Information Processing Systems}, 36:\penalty0 53728--53741, 2023.

\bibitem[Riazanov and Voronkov(2001)]{riazanov2001vampire}
Alexandre Riazanov and Andrei Voronkov.
\newblock Vampire 1.1.
\newblock In \emph{Automated Reasoning: First International Joint Conference, IJCAR 2001 Siena, Italy, June 18--22, 2001 Proceedings 1}, pages 376--380. Springer, 2001.

\bibitem[Schulman et~al.(2017)Schulman, Wolski, Dhariwal, Radford, and Klimov]{schulman2017proximal}
John Schulman, Filip Wolski, Prafulla Dhariwal, Alec Radford, and Oleg Klimov.
\newblock Proximal policy optimization algorithms.
\newblock \emph{arXiv preprint arXiv:1707.06347}, 2017.

\bibitem[Shao et~al.(2024)Shao, Wang, Zhu, Xu, Song, Bi, Zhang, Zhang, Li, Wu, et~al.]{shao2024deepseekmath}
Zhihong Shao, Peiyi Wang, Qihao Zhu, Runxin Xu, Junxiao Song, Xiao Bi, Haowei Zhang, Mingchuan Zhang, YK~Li, Y~Wu, et~al.
\newblock Deepseekmath: Pushing the limits of mathematical reasoning in open language models.
\newblock \emph{arXiv preprint arXiv:2402.03300}, 2024.

\bibitem[Szegedy(2020)]{szegedy2020promising}
Christian Szegedy.
\newblock A promising path towards autoformalization and general artificial intelligence.
\newblock In \emph{Intelligent Computer Mathematics: 13th International Conference, CICM 2020, Bertinoro, Italy, July 26--31, 2020, Proceedings 13}, pages 3--20. Springer, 2020.

\bibitem[Wang et~al.(2023)Wang, Xin, Zheng, Li, Liu, Cao, Huang, Xiong, Shi, Xie, et~al.]{wang2023lego}
Haiming Wang, Huajian Xin, Chuanyang Zheng, Lin Li, Zhengying Liu, Qingxing Cao, Yinya Huang, Jing Xiong, Han Shi, Enze Xie, et~al.
\newblock Lego-prover: Neural theorem proving with growing libraries.
\newblock \emph{arXiv preprint arXiv:2310.00656}, 2023.

\bibitem[Wang et~al.(2025)Wang, Xin, Liu, Li, Huang, Lu, Yang, Tang, Yin, Li, et~al.]{wang2025proving}
Haiming Wang, Huajian Xin, Zhengying Liu, Wenda Li, Yinya Huang, Jianqiao Lu, Zhicheng Yang, Jing Tang, Jian Yin, Zhenguo Li, et~al.
\newblock Proving theorems recursively.
\newblock \emph{Advances in Neural Information Processing Systems}, 37:\penalty0 86720--86748, 2025.

\bibitem[Wang et~al.(2017)Wang, Tang, Wang, and Deng]{wang2017premise}
Mingzhe Wang, Yihe Tang, Jian Wang, and Jia Deng.
\newblock Premise selection for theorem proving by deep graph embedding.
\newblock \emph{Advances in neural information processing systems}, 30, 2017.

\bibitem[Wu et~al.(2022)Wu, Jiang, Li, Rabe, Staats, Jamnik, and Szegedy]{wu2022autoformalization}
Yuhuai Wu, Albert~Qiaochu Jiang, Wenda Li, Markus Rabe, Charles Staats, Mateja Jamnik, and Christian Szegedy.
\newblock Autoformalization with large language models.
\newblock \emph{Advances in Neural Information Processing Systems}, 35:\penalty0 32353--32368, 2022.

\bibitem[Xie et~al.(2025)Xie, chen, Chen, Mao, Xu, and Kong]{xie2025teachinglanguagemodelscritique}
Zhihui Xie, Jie chen, Liyu Chen, Weichao Mao, Jingjing Xu, and Lingpeng Kong.
\newblock Teaching language models to critique via reinforcement learning, 2025.
\newblock URL \url{https://arxiv.org/abs/2502.03492}.

\bibitem[Xin et~al.(2024)Xin, Guo, Shao, Ren, Zhu, Liu, Ruan, Li, and Liang]{xin2024deepseek}
Huajian Xin, Daya Guo, Zhihong Shao, Zhizhou Ren, Qihao Zhu, Bo~Liu, Chong Ruan, Wenda Li, and Xiaodan Liang.
\newblock Deepseek-prover: Advancing theorem proving in llms through large-scale synthetic data.
\newblock \emph{arXiv preprint arXiv:2405.14333}, 2024.

\bibitem[Yang et~al.(2023)Yang, Swope, Gu, Chalamala, Song, Yu, Godil, Prenger, and Anandkumar]{yang2023leandojo}
Kaiyu Yang, Aidan Swope, Alex Gu, Rahul Chalamala, Peiyang Song, Shixing Yu, Saad Godil, Ryan~J Prenger, and Animashree Anandkumar.
\newblock Leandojo: Theorem proving with retrieval-augmented language models.
\newblock \emph{Advances in Neural Information Processing Systems}, 36:\penalty0 21573--21612, 2023.

\bibitem[Zhao et~al.(2024)Zhao, Zheng, Bo, Hu, Thakker, and Kong]{zhao2024subgoalxl}
Xueliang Zhao, Lin Zheng, Haige Bo, Changran Hu, Urmish Thakker, and Lingpeng Kong.
\newblock Subgoalxl: Subgoal-based expert learning for theorem proving.
\newblock \emph{arXiv preprint arXiv:2408.11172}, 2024.

\bibitem[Zheng et~al.(2023)Zheng, Wang, Xie, Liu, Sun, Xin, Shen, Li, and Li]{zheng2023lyra}
Chuanyang Zheng, Haiming Wang, Enze Xie, Zhengying Liu, Jiankai Sun, Huajian Xin, Jianhao Shen, Zhenguo Li, and Yu~Li.
\newblock Lyra: Orchestrating dual correction in automated theorem proving.
\newblock \emph{arXiv preprint arXiv:2309.15806}, 2023.

\bibitem[Zheng et~al.(2021)Zheng, Han, and Polu]{zheng2021minif2f}
Kunhao Zheng, Jesse~Michael Han, and Stanislas Polu.
\newblock Minif2f: a cross-system benchmark for formal olympiad-level mathematics.
\newblock \emph{arXiv preprint arXiv:2109.00110}, 2021.

\end{thebibliography}

\appendix

\section{}\label{apd:first} LLM generated formalization for access policy code. This is a representative example that takes the reader through the different stages of the Proof Generation and verification as show in Figure \ref{fig:approach}.

\begin{lstlisting}[language=json,caption=Problem Name]
{
  "problem_name": "s3_samples_mutations_ec2_exp_single_ec2_prevent_running_classic_policy_6_0"
}
\end{lstlisting}

\begin{lstlisting}[language=json,caption=EC2 Access Policy JSON]
{
  "policy_json": {
    "Statement": [
      {
        "Effect": "Allow",
        "Action": "ec2:RunInstances",
        "Resource": "arn:aws:ec2:us-east-1:123412341234:*"
      },
      {
        "Effect": "Allow",
        "Action": "ec2:RunInstances",
        "Resource": [
          "arn:aws:ec2:us-east-1::image/ami-*",
          "arn:aws:ec2:us-east-1:123412341243:instance/*",
          "arn:aws:ec2:us-east-1:123412341234:volume/*",
          "arn:aws:ec2:us-east-1:123412341234:network-interface/*",
          "arn:aws:ec2:us-east-1:123412341234:key-pair/*"
        ]
      }
    ]
  }
}
\end{lstlisting}

\begin{lstlisting}[language=json,caption=Informal Statement]
{
  "informal_statement": "The text you provided is a policy statement written in JSON format, which is typically used in cloud computing environments like Amazon Web Services (AWS) to define permissions. Here's a breakdown of what it means in plain English:\n\n1. General Permission:\n - The policy allows the action \"ec2:RunInstances.\" This means that the user or service with this policy can start or launch new EC2 instances.\n - This applies to any resource within the specified AWS account (123412341234) in the \"us-east-1\" region.\n\n2. Specific Permissions:\n - Images: use any AMI in \"us-east-1\".\n - Instances: manage EC2 instances in 123412341243.\n - Volumes: manage EBS volumes in 123412341234.\n - Network Interfaces and Key Pairs: full control in 123412341234.\n\nSummary: This policy allows launching and managing EC2 instances and their dependencies."
}
\end{lstlisting}

\begin{lstlisting}[language=json,caption=Informal Proof]
{
  "informal_proof": "To provide an informal proof or argument supporting the interpretation of the JSON policy statement, we break it into structured observations:\n\n1. JSON uses keys like \"Effect\", \"Action\", and \"Resource\" to structure permissions.\n2. 'Allow' means access is granted; 'ec2:RunInstances' lets the user launch EC2s.\n3. '*' in the ARN means it applies to all of that type within the account and region.\n4. Specific ARNs grant permissions to manage AMIs, Instances, Volumes, Interfaces, and Key Pairs.\n\nConclusion: These combined statements demonstrate full EC2 launch and management capability."
}
\end{lstlisting}

\begin{lstlisting}[language=Isar,caption=Formal Statement]
datatype ec2_action = RunInstances

datatype ec2_resource = AllResources | Images | Instances | Volumes | NetworkInterfaces | KeyPairs

datatype principal = Anyone

record policy_entry =
  act :: ec2_action
  res :: ec2_resource
  prin :: principal

definition ec2_instance_policy :: policy_entry where
  "ec2_instance_policy = (|
    act = RunInstances,
    res = AllResources,
    prin = Anyone
  |)"

fun policy_allows :: "policy_entry => ec2_action => ec2_resource => bool" where
  "policy_allows pe a r = (act pe = RunInstances ∧ (res pe = AllResources \/ res pe = r))"

theorem ec2_policy_correctness:
  shows "policy_allows ec2_instance_policy RunInstances AllResources ∧
         policy_allows ec2_instance_policy RunInstances Images ∧
         policy_allows ec2_instance_policy RunInstances Instances ∧
         policy_allows ec2_instance_policy RunInstances Volumes ∧
         policy_allows ec2_instance_policy RunInstances NetworkInterfaces ∧
         policy_allows ec2_instance_policy RunInstances KeyPairs"
  oops
\end{lstlisting}

\begin{lstlisting}[language=Isar,caption=Initial Proof Attempt]
(* Proof of the theorem *)
(*
proof -
  have "policy_allows ec2_instance_policy RunInstances AllResources"
    by (simp add: ec2_instance_policy_def)
  moreover have "policy_allows ec2_instance_policy RunInstances Images"
    by (simp add: ec2_instance_policy_def)
  moreover have "policy_allows ec2_instance_policy RunInstances Instances"
    by (simp add: ec2_instance_policy_def)
  moreover have "policy_allows ec2_instance_policy RunInstances Volumes"
    by (simp add: ec2_instance_policy_def)
  moreover have "policy_allows ec2_instance_policy RunInstances NetworkInterfaces"
    by (simp add: ec2_instance_policy_def)
  moreover have "policy_allows ec2_instance_policy RunInstances KeyPairs"
    by (simp add: ec2_instance_policy_def)
  ultimately show ?thesis by simp
qed
*)
\end{lstlisting}

\begin{lstlisting}[language=Isar,caption=Sorry Proof (Commented)]
(* Proof of the theorem *)
(*
proof -
  have "policy_allows ec2_instance_policy RunInstances AllResources"
    by (simp add: ec2_instance_policy_def)
  moreover have "policy_allows ec2_instance_policy RunInstances Images"
    by (simp add: ec2_instance_policy_def)
  moreover have "policy_allows ec2_instance_policy RunInstances Instances"
    by (simp add: ec2_instance_policy_def)
  moreover have "policy_allows ec2_instance_policy RunInstances Volumes"
    by (simp add: ec2_instance_policy_def)
  moreover have "policy_allows ec2_instance_policy RunInstances NetworkInterfaces"
    by (simp add: ec2_instance_policy_def)
  moreover have "policy_allows ec2_instance_policy RunInstances KeyPairs"
    by (simp add: ec2_instance_policy_def)
  ultimately show ?thesis by simp
qed
*)
\end{lstlisting}

\begin{lstlisting}[language=json,caption=State Information]
{
  "success": true,
  "i_try": 0,
  "success_stage": "init_proof",
  "has_timeout": false,
  "extra_calls": 0,
  "has_sc": false
}
\end{lstlisting}

%\section{Second Appendix}\label{apd:second}This is the second appendix.

\end{document}